\documentclass[10pt,twocolumn,letterpaper]{article}

\usepackage[pagenumbers]{cvpr} 

\def\papername{AMB3R-SLAM}

\usepackage{placeins}
\usepackage[dvipsnames]{xcolor}

\definecolor{firstplace}{rgb}{0.56, 0.78, 0.58}
\definecolor{secondplace}{rgb}{0.9, 1.0, 0.9} 
\definecolor{thirdplace}{rgb}{1.0, 1.0, 0.8}

\definecolor{papercolor}{rgb}{0.988, 0.549, 0.012}

\usepackage{tikz}
\usetikzlibrary{calc}
\usetikzlibrary{backgrounds}

\usepackage{graphicx}
\usepackage{gensymb}
\usepackage{multirow}
\usepackage{array}
\usepackage{makecell}
\usepackage{tabularx}
\usepackage{booktabs}
\usepackage{colortbl}
\usepackage{pifont}

\usepackage{amsmath}
\usepackage{amssymb}

\usepackage{url}

\def\rvu{{\mathbf{i}}}

\def\rvu{{\mathbf{u}}}

\def\gC{{\mathcal{C}}}

\def\gE{{\mathcal{E}}}

\def\gM{{\mathcal{M}}}

\def\gV{{\mathcal{V}}}

\definecolor{cvprblue}{rgb}{0.21,0.49,0.74}
\usepackage[pagebackref,breaklinks,colorlinks,allcolors=cvprblue]{hyperref}

\def\paperID{111}
\def\confName{3DV\xspace}
\def\confYear{2027\xspace}

\title{AMB3R-SLAM: Kilometer-scale SLAM with Hierarchical Backend}

\author{Hengyi Wang \quad Lourdes Agapito\\[2.5pt]
\vspace{2pt}
Department of Computer Science, University College London\\
\vspace{1pt}
{\tt\small \url{https://hengyiwang.github.io/projects/amber-slam}
}}

\begin{document}
\twocolumn[{%
    \renewcommand\twocolumn[1][]{#1}%
    \maketitle
    \centering
    \vspace{-0.6cm}
    \vspace{0mm}
\includegraphics[width=0.95\linewidth]{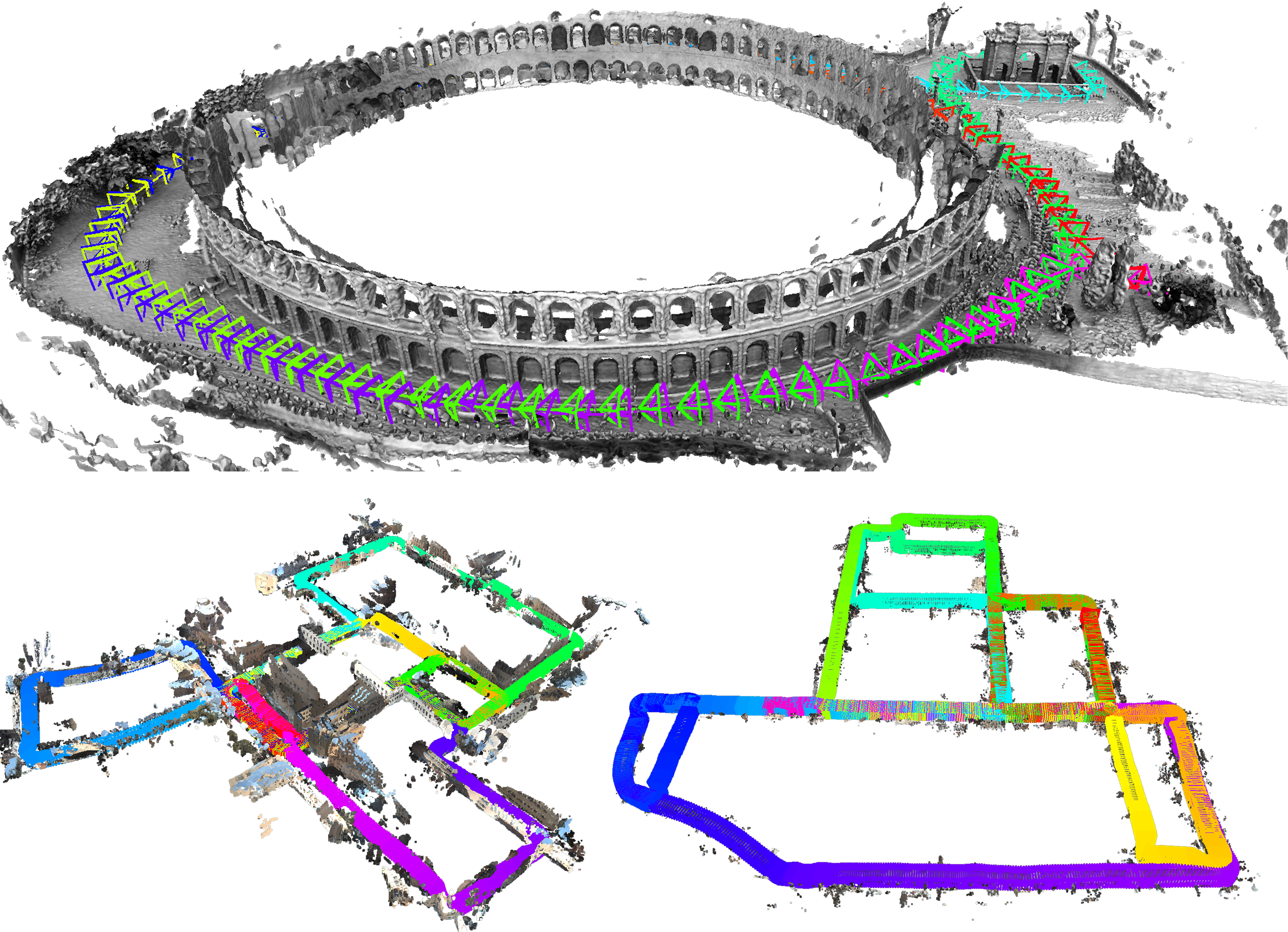}
\vspace{0mm}
\captionof{figure}{\textbf{Overview.} \papername{} is a real-time monocular SLAM system that can reconstruct kilometer-scale trajectories over 10k frames on a single consumer-grade GPU. AMB3R-SLAM can handle both static and dynamic scenes and can be extended to take stereo, RGB-D, or LiDAR as the additional input. }
    \vspace{0.6cm}
}]
\begin{abstract}
We present AMB3R-SLAM, a real-time monocular SLAM system capable of reconstructing kilometer-scale trajectories over 10k frames on a single consumer-grade GPU. Our model couples a lightweight front-end for low-latency online tracking with a hierarchical backend that progressively enforces local, mid-level, and global consistency. By avoiding bundle adjustment that relies on the static world assumption, our system naturally handles complex dynamic scenes out of the box. \enlargethispage{\baselineskip}Furthermore, we demonstrate that our method can be extended to leverage stereo, RGB-D, and LiDAR as additional inputs. AMB3R-SLAM achieves strong camera tracking performance across 9 datasets, reducing the absolute trajectory error (ATE) of previous state-of-the-art methods on VBR and Oxford Spires by over 70\%. With additional LiDAR input, our model further reduces ATE to sub-meter level on KITTI and VBR datasets. 
\end{abstract}
\section{Introduction}
\label{sec:intro}

Visual Simultaneous Localization and Mapping (SLAM) aims to perform online camera tracking and dense 3D reconstruction from a video stream. Recently, 3D geometric foundation models~\cite{wang2024dust3r,wang2025vggt,lin2025depthanything3} have shown remarkable capabilities in end-to-end dense reconstruction. Given dozens of images at once, they leverage a transformer that attends to all views to directly regress depthmaps and camera poses in an offline manner. However, visual SLAM requires a continuous streaming rollout with low latency and bounded memory. This makes it challenging to adapt 3D foundation models with quadratic complexity to this task.

Recent attempts to extend geometric foundation models to streaming video typically leverage causal attention or recursive memory compression~\cite{wang2025spann3r,wang2025cut3r,zhang2026loger,chen2026lingbotmap} instead of dense bidirectional attention. While achieving constant per-frame runtime and bounded memory, these approaches sacrifice local reconstruction accuracy, and the sequential chain of causal methods inevitably leads to drift. In contrast, AMB3R-VO~\cite{wang2026amb3r} proposes a model-agnostic visual odometry pipeline that resolves the trade-off between streaming efficiency and geometric fidelity by retaining full bidirectional attention within a compact set of active keyframes. However, AMB3R-VO remains a visual odometry pipeline; without global optimization and long-range constraints, it still accumulates drift over long trajectories.

To bridge the gap between feed-forward visual odometry and large-scale SLAM, we present \papername{}, a real-time monocular SLAM system that can reconstruct kilometer-scale trajectories over 10k frames on a single consumer-grade GPU. Compared to existing optimization-based SLAM systems with dense reconstruction priors~\cite{murai2025mast3rslam,zhang2025vista,maggio2025vggtslam}, our design principle is to rely on the feed-forward predictions of geometric foundation models and avoid optimization that relies on post-hoc estimation (e.g., correspondence). In this way, our system can inherit the rich geometric priors learned from massive data, remaining robust across various challenging conditions. 

\papername{} consists of a front-end that performs low-latency per-frame camera tracking and a hierarchical backend that progressively ensures local, mid-level, and global consistency via dense local submapping with span-2 edges, sparse long-context mapping, and loop closure. By avoiding bundle adjustment that only works with a static world, our method naturally handles complex dynamic scenes out of the box. Furthermore, we demonstrate that our method can be easily extended to stereo, RGB-D, and LiDAR cameras. Our experimental results across 9 datasets show that \papername{} achieves strong camera tracking results and reduces the absolute trajectory error of previous state-of-the-art methods on VBR~\cite{brizi2024vbr} and Oxford Spires~\cite{tao2025oxfordspires} datasets by over 70\%. With LiDAR as additional input, we can further reduce the tracking error on the VBR~\cite{brizi2024vbr} and KITTI~\cite{geiger2012kitti} datasets to the sub-meter level.

\section{Related Work}
\label{sec:related_work}

\subsection{Geometric Foundation Models}

Geometric foundation models~\cite{wang2024dust3r, jang2025pow3r, leroy2024mast3r,wang2025vggt,lin2025depthanything3} aim to predict dense geometry and camera pose in a feed-forward manner following either an incremental paradigm~\cite{wang2025spann3r,cabon2025must3r,wang2025cut3r,liu2025slam3r,zhuo2025streamvggt,chen2025long3r,lan2025stream3r,li2025wint3r,wu2025point3r,ma2025puzzles,khafizov2025g,mahdi2025evict3r,chen2025ttt3r,yuan2025slam,wang2025pi3} or a global paradigm~\cite{wang2025vggt,elflein2025light3rsfm,yang2025fast3r,tang2025mvdust3r,wang2025pi3,wang2025fastervggt,gao2026more,xu2026r3,keetha2025mapanything,wang2026amb3r,fang2025dens3r}. The learned data-driven prior enables unprecedented generalization capability as well as robustness towards various challenging conditions. This also allows dynamic scene reconstruction~\cite{jiang2025geo4d,fei2024driv3r,team2025aether,mai2025can,li2025stereodiff,zhang2025pomato,yu2026vge} by learning from diverse video data. For large-scale scene reconstruction, several works~\cite{deng2025vggtlong,kani2026g3t,xie2026scal3r} divide video into overlapping chunks and map them together with global optimization. However, these methods, despite being able to perform video mapping, are chunk-based offline methods by construction. Several other methods~\cite{zhang2026loger,wang2026amb3r,chen2026lingbotmap,yugay2026calfvo} instead leverage geometric foundation models themselves to perform visual odometry. However, these approaches tend to drift due to the lack of global consistency.

\subsection{Visual Odometry and SLAM}
\label{sec:related_slam}

Visual Odometry (VO)~\cite{nister2004vo,engel2017dso,forster2014svo,teed2023dpvo,wang2026amb3r} aims to estimate the 6-DoF camera ego-motion between consecutive frames to track local trajectories in real time, while visual SLAM~\cite{davison2007monoslam,klein2007ptam,mur2015orb} concurrently constructs a globally consistent 3D map via loop closure and global optimization to correct cumulative drift. Traditional visual SLAM is broadly categorized into feature-based methods~\cite{rublee2011orb,mur2015orb} and direct methods~\cite{engel2014lsd,engel2017dso}. Beyond monocular SLAM, multi-modal SLAM can integrate depth~\cite{newcombe2011kinectfusion,whelan2015elasticfusion}, stereo~\cite{mur2017orbslam2}, or LiDAR sensors~\cite{zhang2014loam,shan2020liosam,vizzo2023kissicp,shan2021lvisam} to provide robust metric-scale camera trajectory estimation. Later, research shifted towards leveraging learned dense correspondences and differentiable bundle adjustment~\cite{teed2021droidslam,teed2023dpvo,lipson2024dpvslam}, neural implicit representations~\cite{sucar2021imap,zhu2022nice,wang2023coslam,kong2023vmap,pan2024pinslam}, or 3D Gaussian Splatting~\cite{matsuki2024monogs,yugay2023gaussianslam,huang2024photo} for SLAM. Recently, advances in geometric foundation models~\cite{leroy2024mast3r,wang2025vggt} have enabled SLAM to operate in uncalibrated settings~\cite{murai2025mast3rslam,maggio2025vggtslam,zhang2025vista}. However, these works still struggle to scale to long-horizon, large-scale environments. In this work, we push the frontier of uncalibrated visual SLAM to kilometer-scale sequences via a hierarchical backend that progressively ensures local, mid-level, and global consistency. Our system can also be extended to take stereo, RGB-D, and LiDAR as additional inputs for metric-scale camera tracking and reconstruction.

\begin{figure*}[t]
    \centering
    \includegraphics[width=\textwidth]{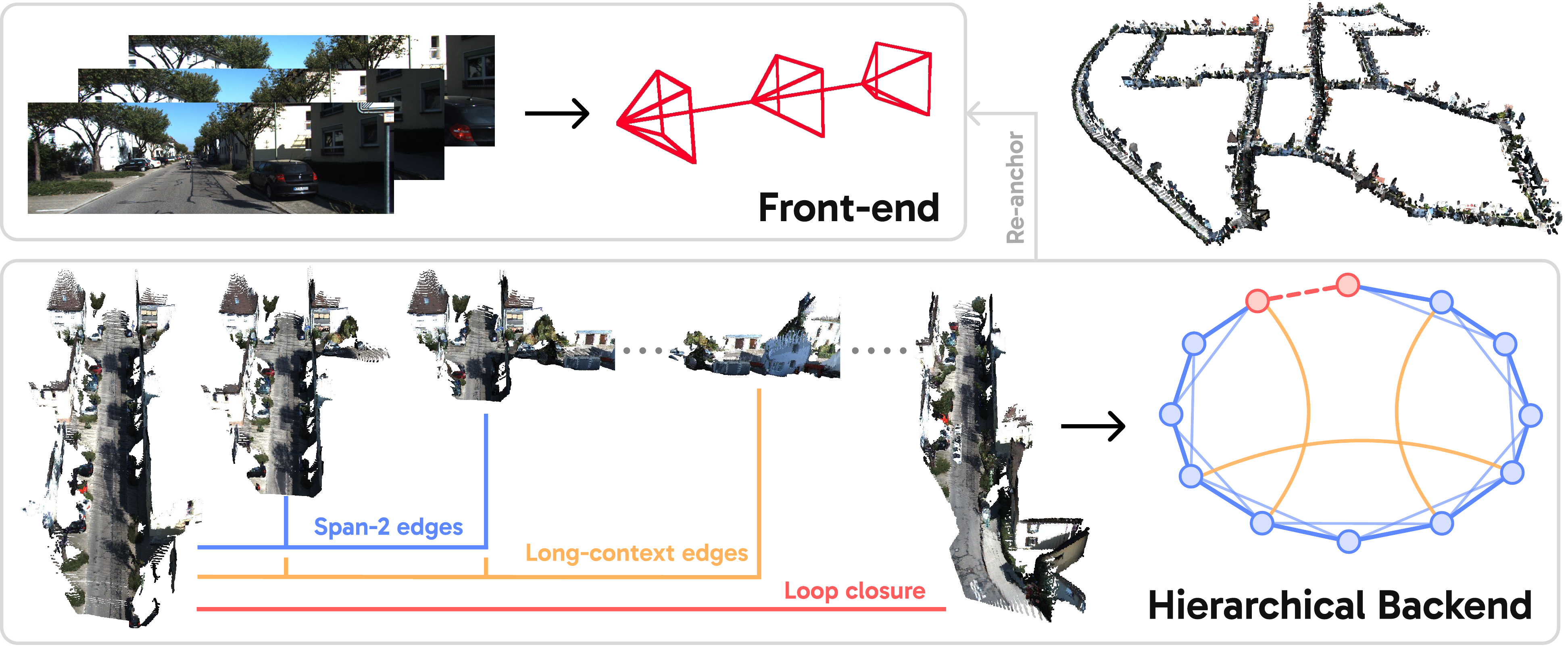}
     \vspace{-4mm}
    \caption{\textbf{Overview of \papername{}.} \papername{} consists of a front-end that performs low-latency camera pose tracking and a hierarchical backend that progressively ensures local, mid-level, and global consistency through pose graph optimization with span-2, long-context, and loop closure edges. } 
    \label{fig:overview}
    \vspace{-1mm}
\end{figure*} 
\section{Method}
\label{sec:method}

Fig.~\ref{fig:overview} shows an overview of our method. \papername{} couples a lightweight front-end for low-latency camera tracking with a hierarchical backend that progressively enforces local, mid-level, and global consistency: overlapping submaps provide accurate local reconstruction and drift correction via Span-2 edges, sparse long-context mapping introduces mid-range constraints, and loop closures enforce global consistency. We further show that \papername{} naturally handles dynamic scenes without task-specific adaptations and can incorporate additional modalities, including stereo, RGB-D, and LiDAR.

\subsection{Front-end}
\label{sec:frontend}

The front-end performs low-latency online camera pose tracking of each new frame within the current submap. To maintain a high frame rate, we employ a lightweight backbone, i.e., DA3-Small~\cite{lin2025depthanything3} with 80M parameters for our front-end. For each new frame, we construct a compact memory with the anchor keyframe and the two most recent frames to estimate the camera poses and solve the scale factor via a robust scale solver as in AMB3R-VO~\cite{wang2026amb3r}. We show its efficiency and accuracy in Tab.~\ref{tab:front_end}. The confidence of the front-end is used for view selection in the backend, and we constantly re-anchor the front-end via backend estimation to reset accumulated local drift.

\begin{table}[t!]
\centering
\setlength{\tabcolsep}{2pt}
\caption{\textbf{Evaluation on Front-end drift and efficiency.} Front-end drift is evaluated within each mapping window before re-anchoring. Runtime (FPS) and peak memory usage are reported separately per dataset due to resolution differences.}
\vspace{-2mm}
\resizebox{\columnwidth}{!}{
\begin{tabular}{l cccc cccc}
\toprule
\multirow{2}{*}{\textbf{Method}} 
& \multicolumn{4}{c}{\textbf{KITTI}~\cite{geiger2012kitti}} 
& \multicolumn{4}{c}{\textbf{TUM}~\cite{sturm2012tumrgbd}} 
\\ \cmidrule(lr){2-5} \cmidrule(lr){6-9}
& $R_{\text{err}}\!\downarrow$ & $\text{ATE}\%\!\downarrow$ & FPS & Mem [GB]
& $R_{\text{err}}\!\downarrow$ & $\text{ATE}\%\!\downarrow$ & FPS & Mem [GB]
\\ \midrule

Two-view & 
2.61 & 12.09 & \textbf{105.6} & \textbf{0.30} & 
5.87 & 8.84  & \textbf{70.3} & \textbf{0.45} \\

Ours (S) &
1.41 & 4.43 & 78.3 & 0.40 &
2.78 & 5.08 & 44.8 & 0.70 \\

Ours (B) & 
\textbf{1.40} & \textbf{2.87} & 48.0 & 1.12 & 
\textbf{1.60} & \textbf{4.33} & 22.9 & 1.72 \\

\bottomrule
\end{tabular}}

\label{tab:front_end}
\vspace{-1mm}
\end{table}

\subsection{Hierarchical Backend}
\label{sec:backend}

\noindent
\textbf{Dense Local Mapping.} While the front-end provides low-latency tracking, its compact context inevitably exhibits tracking drift over time. To leverage multi-view information over sufficiently wide baselines, the backend constructs dense submaps $\gM_k$ every $\Delta$ frames over a sliding window of $n$ consecutive frames. We set $\Delta = n/3$, which ensures that each submap overlaps not only with its neighbor $\gM_{k+1}$ but also with $\gM_{k+2}$, naturally enabling Span-2 edges. To limit backend computation cost, we apply uniform temporal grouping with stride $d_{\text{stride}}$ within each submap. The backend reconstructs only the frames with the highest confidence in each group and interpolates intermediate poses, enabling the model to perceive wide baselines with a small number of context frames. Here, we deploy a large foundation model~\cite{lin2025depthanything3,wang2026vggtomega} and set the middle frame of each submap as the reference view to reduce pose drift at the submap boundaries. Coordinate alignment across overlapping submaps follows AMB3R-VO~\cite{wang2026amb3r}.

\noindent
\textbf{Sparse Long-Context Mapping.}
Local submaps are only connected to their Span-2 neighbors. This sequential structure inevitably accumulates drift over long trajectories without physical loop revisits. To suppress drift without relying on loop closures, we perform an additional sparse long-context mapping over multiple submaps to supply long-range constraints for the pose graph. Every $M$ submaps, we construct a cross-submap window covering $K$ consecutive submaps. To limit computational cost, we sample a sparse set of keyframes across these $K$ submaps to produce an intermediate reconstruction $\gC$. From $\gC$, we extract similarity transformations $G_{ij} \in \mathrm{Sim}(3)$ between distant submaps $\gM_i$ and $\gM_j$ ($|i - j| \ge 3$) to inject long-span edges into the pose graph. Because the model might degrade under long-context mapping with a sparse set of keyframes, we admit an edge $G_{ij}$ into the graph only if: (i) the confidence of $\gC$ remains within a fraction of the local submaps; and (ii) the covisibility between $\gM_i$ and $\gM_j$ is sufficient.

\noindent
\textbf{Loop Closure.} 
Although sparse long-context mapping mitigates drift, kilometer-scale trajectories still require explicit loop closures to enforce global consistency. We employ DBoW2~\cite{galvez2012bags} to propose loop candidates between a historical submap $\gM_i$ and the current submap $\gM_j$. For each candidate, we sample views from both submaps and jointly reconstruct them with our foundation model, computing 3D voxel occupancy overlap for local geometric verification. We filter out loop edges with implausible rotation and translation corrections to prevent perceptual aliasing in repetitive environments. Verified loop edges then enter the pose graph optimization in Eq.~\ref{eq:pgo}.

\begin{figure*}[t]
    \centering
    \includegraphics[width=0.95\textwidth]{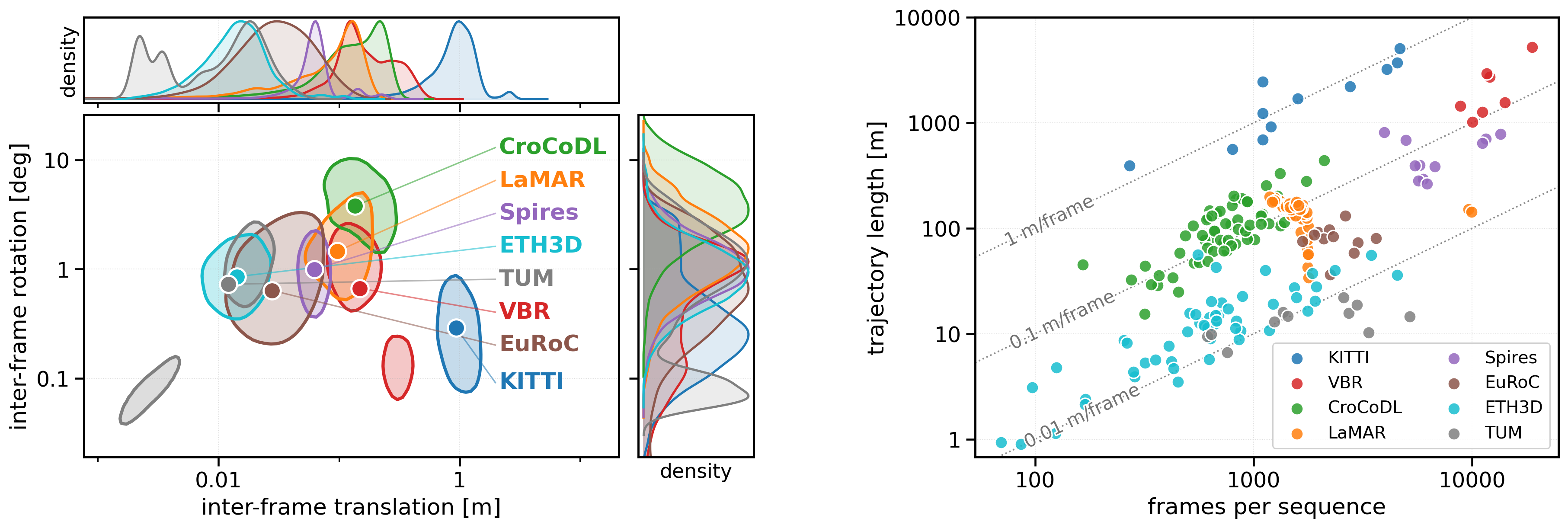}
     \vspace{-2mm}
    \caption{\textbf{Statistics of evaluation dataset.} We visualize the statistics of rotation and translation between adjacent frames, as well as the frame count and trajectory length per sequence. For LaMAR, we calibrate the statistics by their gt frame stride. } 
    \label{fig:data}
    \vspace{-1mm}
\end{figure*}

\noindent
\textbf{Pose Graph Optimization.} We construct a $\mathrm{Sim}(3)$ pose graph over keyframe nodes $\{S_i \in \mathrm{Sim}(3)\}$ for global trajectory optimization. The objective over the Span-2, long-span, and loop closure edges $(i,j) \in \gE$ is formulated as:
\begin{equation}
  \{S_i^\star\} = \arg\min_{\{S_i\}} \sum_{(i,j) \in \gE} 
  \rho\Big( \mathbf{r}_{ij}^T \mathbf{W}_{ij}^T \mathbf{W}_{ij} \mathbf{r}_{ij} \Big),
  \label{eq:pgo}
\end{equation}
where $\mathbf{r}_{ij} = \log_{\mathrm{Sim}(3)}\big( G_{ij}^{-1} S_i^{-1} S_j \big) = [\boldsymbol{\nu}_{ij}^T, \boldsymbol{\phi}_{ij}^T, \sigma_{ij}]^T \in \mathfrak{sim}(3) \cong \mathbb{R}^7$ decomposes the error into translation ($\boldsymbol{\nu} \in \mathbb{R}^3$), rotation ($\boldsymbol{\phi} \in \mathbb{R}^3$), and log-scale ($\sigma \in \mathbb{R}$). To balance translation, rotation, and scale across varying distances, we define the whitening matrix as:
\begin{equation}
  \mathbf{W}_{ij} = \sqrt{w_{ij}} \operatorname{diag}\left( 
    \frac{\mathbf{I}_3}{d_{ij}}, \;
    \mathbf{I}_3, \;
    1
  \right),
  \label{eq:whitening}
\end{equation}
where $w_{ij}$ is the edge weight, and $d_{ij} = \max(\|\mathbf{t}_{ij}\|, d_{\min})$ scales the translation error by the edge's baseline, with a small floor $d_{\min}$ avoiding division by zero. This measures relative rather than absolute drift, preventing long-range translation error from dominating optimization. We adopt a Huber loss $\rho$ to suppress outliers. The resulting $\mathrm{Sim}(3)$ corrections are propagated to all frames, using weighted pose averaging~\cite{wang2026amb3r} on overlapping windows.

\subsection{Robustness to Dynamic Scenes}
\label{sec:dynamic}

Traditional visual SLAM systems heavily rely on bundle adjustment, which inherently assumes a static world and easily breaks in the presence of dynamic objects. In contrast, our framework eliminates static-scene assumptions by design. First, the feed-forward foundation models~\cite{lin2025depthanything3,wang2026vggtomega} are trained on diverse dynamic data, naturally learning to disentangle rigid camera ego-motion from rigid or non-rigid motion of moving objects. Second, our hierarchical backend optimizes a $\mathrm{Sim}(3)$ pose graph strictly over relative camera poses rather than 2D--3D reprojection errors across moving pixels. Consequently, our system handles complex dynamic environments out of the box, without requiring explicit motion segmentation.

\subsection{Extension to Stereo and RGB-D}
\label{sec:modalities}
Monocular SLAM suffers from scale ambiguity, leading to cumulative drift across sequential submaps in the absence of loop closures. Given stereo or RGB-D inputs, we constrain this scale drift by calibrating submaps to metric scale. We obtain metric depth $d_t^{\mathrm{s}}$ directly from RGB-D sensors or via semi-global matching~\cite{hirschmuller2008sgm} for calibrated stereo pairs. For each submap $\gM_k$, we estimate a metric scale factor $\hat{s}_k$ via:
\begin{equation}
  \hat{s}_k = \operatorname{median}_{t \in \gM_k, \,\rvu \in \gV_{k,t}} 
  \left( \frac{d_t^{\mathrm{s}}(\rvu)}{d_{k,t}(\rvu)} \right),
  \label{eq:metric_scale}
\end{equation}
\noindent
where $\gV_{k,t}$ denotes valid pixels. We discard scale estimates with a low depth inlier ratio to avoid non-linear shape distortions. Once submaps are calibrated to metric scale, their relative scale is fixed to $s_{ij}=1.0$. Consequently, the pose graph optimization in Eq.~\ref{eq:pgo} reduces from $\mathrm{Sim}(3)$ to $\mathrm{SE}(3)$, preventing cumulative scale drift along the trajectory.

\subsection{Extension to LiDAR}
\label{sec:lidar_proposal}
Unlike stereo or structured-light sensors, whose depth error grows as $O(z^2)$ with distance through baseline triangulation and signal attenuation, a LiDAR measures range by direct pulsed time-of-flight, with an almost range-invariant $O(1)$ error at centimeter accuracy over hundreds of meters. When LiDAR is available, we instead leverage an ICP-based method~\cite{vizzo2023kissicp} to obtain relative poses in the pose graph. When a loop is detected, we exploit the robustness of our geometric foundation model to estimate the initial loop edge, followed by ICP-based point cloud refinement. Since each loop edge might have different registration quality, we leverage its inlier root-mean-square error (RMSE) $\varepsilon_{ij}$ to weight the loop edges:
\begin{equation}
  w_{ij} = \bar{w} \left( \frac{\bar{\varepsilon}}{\varepsilon_{ij}} \right)^{\!2},
  \label{eq:invvar}
\end{equation}
\noindent
where $\bar{w}$ is the loop edge weight, $\varepsilon_{ij}$ is the inlier RMSE of ICP registration, and $\bar{\varepsilon}$ is the median RMSE across all candidate closures. In this way, high-quality loop edges are assigned larger weights, while potential misalignments are suppressed to ensure robust pose graph optimization.

\section{Experiments}

\subsection{Setup}

\begin{table*}[htbp]
\centering
\colorlet{secolor}{lightgray}
\newcommand{\se}{\color{secolor}}

\setlength{\tabcolsep}{3pt}
\caption{\textbf{Comparison of ATE [m]$\downarrow$ on KITTI.}}
\vspace{-1mm}
\label{tab:kitti_pose}
\resizebox{\textwidth}{!}{
\begin{tabular}{llcccccccccccc}
\toprule
\textbf{Category} & \textbf{Method} 
& {00} & {01} & {02} & {03} & {04} & {05} & {06} & {07} & {08} & {09} & {10} & {Avg.} \\
&  
& \small 4.5k/3.7km & \small 1.1k/2.5km & \small 4.7k/5.1km & \small 0.8k/0.6km & \small 0.3k/0.4km 
& \small 2.8k/2.2km & \small 1.1k/1.2km & \small 1.1k/0.7km & \small 4.1k/3.2km 
& \small 1.6k/1.7km & \small 1.2k/0.9km & \\

\midrule
\multirow{2}{*}{Offline}
& Scal3R \citep{xie2026scal3r}
& 4.30 & 45.29 & 42.06 & 3.36 & 1.74 & 3.30 & 2.49 & 2.03 & 36.69 & 12.32 & 6.46 & 14.55 \\
& Glob3R~\cite{deng2026glob3r} 
& 2.77 & 53.12 & 30.10 & 1.53 & 0.75 & 2.80 
& 2.67 & 1.77 & 32.76 & 5.51 & 11.51 & 13.21 \\

\midrule
\multirow{3}{*}{Chunk}
& VGGT-Long \citep{deng2025vggtlong}
& 8.64 & 61.21 & 52.72 & 8.78 & 4.20 & 9.88 & 4.67 & 2.66 & 72.98 & 31.84 & 27.71 & 25.94 \\
& $\pi$3-Long~\cite{wang2025pi3}
& 5.55 & 114.83 & 50.29 & 1.63 & 1.11 & 3.48 & 2.88 & 3.92 & 24.25 & 7.38 & 17.61 & 21.17 \\
& DA3-Long~\cite{lin2025depthanything3}
& 5.13 & 78.76 & 35.64 & 5.38 & 3.18 & 3.04 & 2.83 & 2.32 & 26.55 & 8.86 & 13.42 & 16.83 \\

\midrule
\multirow{9}{*}{Online}
& DROID-SLAM \citep{teed2021droidslam}
& 92.10 & 344.60 & 107.61 & 2.38 & 1.00 & 118.50 & 62.47 & 21.78 & 161.60 & 72.32 & 118.70 & 100.28 \\
& VGGT-SLAM2 \citep{maggio2026vggtslam2}
& TL & 163.65 & TL & 50.04 & 19.38 & 159.58 & 46.35 & 57.80 & TL & 167.96 & 76.99 & 92.72 \\
& DA3-SLAM
& 91.07 & 263.25 & 136.08 & 3.72 & 6.79 & 50.41 & 115.39 & 33.34 & 28.67 & 67.67 & 14.36 & 73.71 \\
& MASt3R-SLAM \citep{murai2025mast3rslam}
& OOM & 530.37 & OOM & 18.87 & 88.99 & 159.43 & 92.00 & OOM & 263.75 & TL & 153.07 & 186.64 \\
& AMB3R-VO~\cite{wang2026amb3r} 
& 167.38 & 276.12 & 157.07 & 22.06 & 6.91 & 149.11 & 55.94 & 37.25 & 85.68 & 134.73 & 52.15 & 104.04 \\
& LingBot-Map \citep{chen2026lingbotmap}
& 27.17 & 70.94 & 112.02 & 2.02 & 1.36 & 26.14 & 16.61 & 10.48 & 23.82 & 17.84 & 6.48 & 28.63 \\
& LoGeR \cite{zhang2026loger}
& 30.47 & 47.91 & 36.32 & 5.38 & 1.95 & 26.34 & 6.60 & 5.55 & 24.41 & 10.12 & 10.11 & 18.65 \\
& \textbf{Ours}
& 3.67 & 55.91 & 31.55 & 4.80 & 2.04 & 2.68 & 3.33 & 1.73 & 23.56 & 9.03 & 5.96 & 13.11 \\
& \textbf{Ours (Stereo)}
& 2.67 & 56.04 & 29.89 & 2.29 & 0.74 & 1.71 & 3.69 & 1.46 & 23.46 & 5.63 & 6.05 & 12.15 \\
\midrule
\multirow{4}{*}{LiDAR}
& PIN-SLAM~\cite{pan2024pinslam}
& 0.87 & 5.23 & 2.34 & 0.44 & \textbf{0.11} & \textbf{0.26} & \textbf{0.15} & \textbf{0.27} & \textbf{1.85} & 1.20 & \textbf{0.71} & 1.22 \\
& \quad \se$\hookrightarrow$ $\mathrm{SE}(3)$
& \se 0.88 & \se 5.43 & \se 2.71 & \se 0.72 & \se\textbf{0.12} & \se\textbf{0.31} & \se 0.45 & \se\textbf{0.30} & \se\textbf{1.88} & \se 1.23 & \se\textbf{0.74} & \se 1.34 \\
& \textbf{Ours (LiDAR)}
& \textbf{0.86} & \textbf{2.95} & \textbf{1.53} & \textbf{0.27} & 0.16 & 0.28 & 0.19 & 0.28 & 2.03 & \textbf{1.12} & 0.74 & \textbf{0.95} \\
& \quad \se$\hookrightarrow$ $\mathrm{SE}(3)$
& \se\textbf{0.86} & \se\textbf{3.07} & \se\textbf{1.70} & \se\textbf{0.45} & \se 0.28 & \se\textbf{0.31} & \se\textbf{0.29} & \se\textbf{0.30} & \se 2.03 & \se\textbf{1.13} & \se 0.76 & \se\textbf{1.01} \\
\bottomrule
\end{tabular}
}
\vspace{-1mm}
\end{table*}

\begin{table*}[t]
\centering
\setlength{\tabcolsep}{5pt}
\caption{\textbf{Comparison of ATE [m]$\downarrow$ on Oxford Spires.}}
\vspace{-2mm}
\resizebox{0.9\textwidth}{!}{%
\begin{tabular}{l ccccccccccc c}
\toprule
\multirow{2}{*}{\textbf{Methods}}
& \textbf{keb\_02} & \textbf{keb\_03} & \textbf{keb\_04} & \textbf{keb\_05} & \textbf{obs\_01} & \textbf{obs\_02} & \textbf{ble\_05} & \textbf{chr\_02} & \textbf{chr\_03} & \textbf{chr\_05} & \textbf{bod\_02} & \multirow{2}{*}{\textbf{Avg.}} \\
& \scriptsize 6.0k/292m & \scriptsize 5.7k/282m & \scriptsize 13.5k/779m & \scriptsize 11.5k/701m & \scriptsize 5.7k/396m & \scriptsize 5.5k/391m & \scriptsize 6.8k/384m & \scriptsize 11.2k/637m & \scriptsize 6.2k/262m & \scriptsize 4.0k/801m & \scriptsize 5.0k/685m & \\
\midrule
VGGT-SLAM2~\cite{maggio2026vggtslam2} & 20.56 & 16.20 & 24.71 & 30.02 & 8.97 & 7.42 & \textbf{3.56} & 30.09 & 4.90 & 28.13 & 63.24 & 21.62 \\
DA3-SLAM                             & 2.86  & 3.88  & 4.34  & 7.97  & 18.69 & 3.87 & 14.84         & 8.89  & 6.93 & 15.48 & 44.78 & 12.05 \\
LoGeR~\cite{zhang2026loger}          & 6.14 & 4.50 & 5.35 & 13.37 & 7.35 & 8.76 & \underline{5.51} & 9.33 & 1.46 & 21.79 & 10.92 & 8.59 \\
LingBot-Map~\cite{chen2026lingbotmap}  & 4.30 & 2.15 & 4.37 & 6.65 & 4.33 & 3.53 & 6.99 & 13.34 & 2.20 & 36.05 & 10.91 & 8.62 \\
\textbf{Ours ($\Omega$)}             & \textbf{0.75} & \textbf{0.34} & \textbf{2.03} & \textbf{0.74} & \textbf{0.40} & \textbf{0.75} & 5.60 & \textbf{0.72} & \textbf{0.19} & \textbf{1.27} & \underline{10.15} & \textbf{2.08} \\
\textbf{Ours}                        & \underline{0.80} & \underline{0.55} & \underline{2.06} & \underline{1.15} & \underline{1.78} & \underline{1.74} & 6.74 & \underline{2.23} & \underline{0.25} & \underline{2.09} & \textbf{6.11} & \underline{2.32} \\
\bottomrule
\end{tabular}%
}
\label{tbl:spires}
\vspace{-1mm}
\end{table*}

\begin{figure*}[t]
    \centering
    \includegraphics[width=\textwidth]{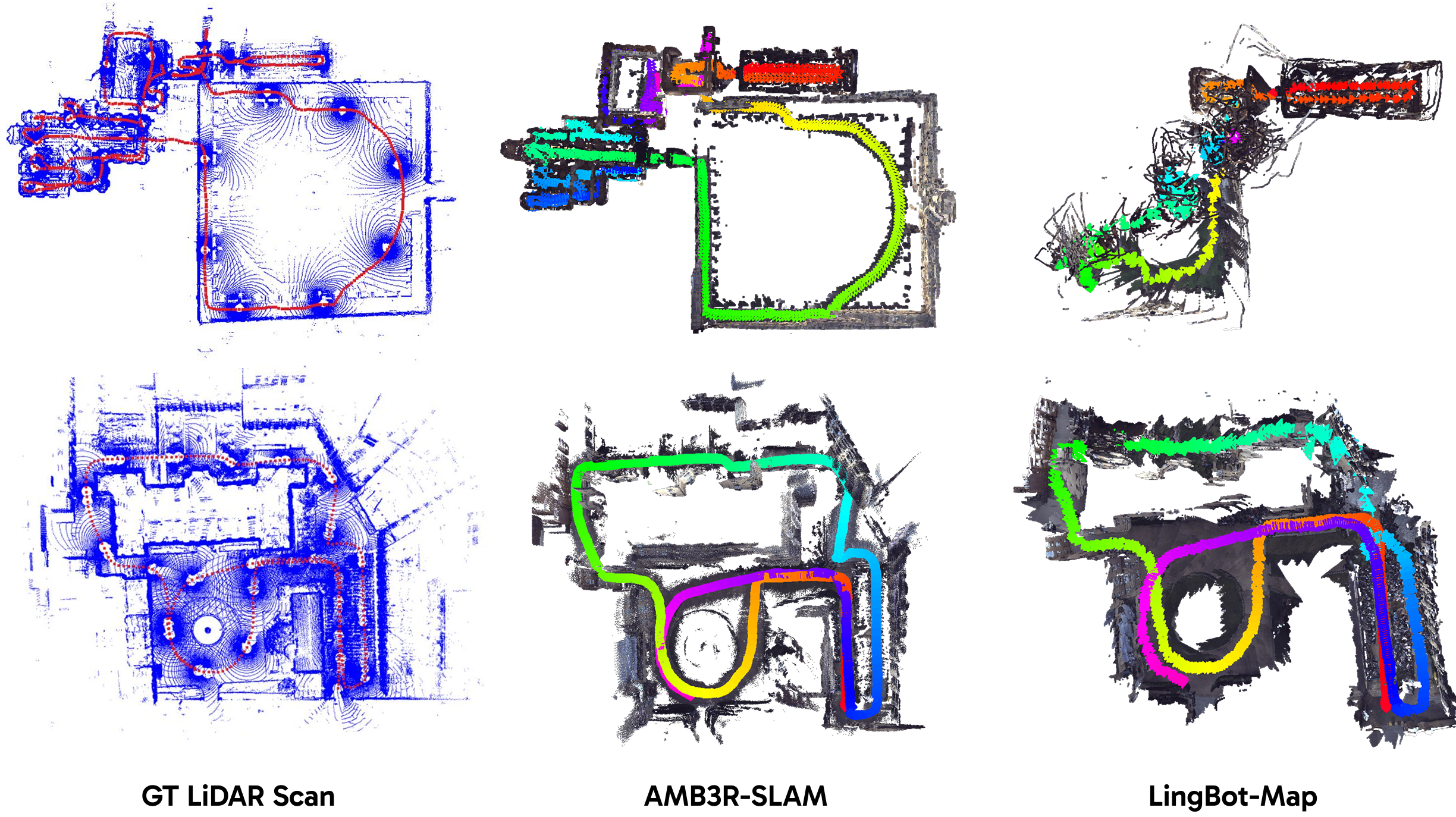}
     \vspace{-4mm}
     \caption{\textbf{Qualitative Showcase on Oxford Spires dataset.}} 
     \vspace{-1mm}
    \label{fig:qual}
\end{figure*}

\noindent
\textbf{Benchmarks.} We evaluate \papername{} on 9 datasets with their trajectory statistics shown in Fig.~\ref{fig:data}. For \emph{kilometer-scale outdoor} tracking, we use KITTI~\cite{geiger2012kitti} (0.3k--4.7k frames, 0.4--5.1\,km), VBR~\cite{brizi2024vbr} (8.8k--18.8k frames, 1.0--5.2\,km), and Oxford Spires~\cite{tao2025oxfordspires} (4.0k--13.5k frames, 0.26--0.80\,km). For \emph{building-scale} captures with mixed indoor\&outdoor traversal, we use LaMAR~\cite{sarlin2022lamar} and CroCoDL~\cite{Blum2025crocodl}. For \emph{room-scale} sequences, we use TUM RGB-D~\cite{sturm2012tumrgbd}, ETH3D-SLAM~\cite{schops2019badslam}, and EuRoC~\cite{Burri2016euroc}. For \emph{dynamic} scenes, we use the Bonn dynamic dataset~\cite{palazzolo2019bonn}. Together, these datasets range from tabletop motions of a few meters to 5\,km of continuous urban driving.

\noindent
\textbf{Baselines.} We compare primarily against state-of-the-art streaming VO/SLAM systems: LoGeR~\cite{zhang2026loger}, LingBot-Map~\cite{chen2026lingbotmap}, MASt3R-SLAM~\cite{murai2025mast3rslam}, VGGT-SLAM2~\cite{maggio2026vggtslam2}, and AMB3R-VO~\cite{wang2026amb3r}. The offline methods~\cite{pan2024glomap,xie2026scal3r,deng2026glob3r,deng2025sail,pataki2026vidmap}, including chunk-based methods~\cite{deng2025vggtlong,wang2025pi3,lin2025depthanything3}, are shown for reference. On dynamic scenes, we additionally compare to methods designed for moving objects~\cite{zheng2025wildgsslam,li2025megasam}. To separate the contribution of our system from that of its backbone, we further integrate DepthAnything~3~\cite{lin2025depthanything3} into the VGGT-SLAM pipeline and refer to this controlled baseline as DA3-SLAM.

\noindent
\textbf{Metrics.} We report absolute trajectory error (ATE RMSE) on KITTI, VBR, Oxford Spires, TUM, and Bonn, following prior work~\cite{murai2025mast3rslam,wang2026amb3r,zhang2026loger,chen2026lingbotmap,zheng2025wildgsslam}. Following VidMap~\cite{pataki2026vidmap}, we report the AUC of the translation error over thresholds from 5\,cm to 10\,m on ETH3D-SLAM and EuRoC. On LaMAR and CroCoDL, and in our ablation on VBR, we report the windowed AUC (W-AUC) of VidMap~\cite{pataki2026vidmap}, which evaluates trajectory segments of a given length against a threshold set to 5\% of that length.

\noindent
\textbf{Implementation details.} Our front-end uses DA3-Small (0.08B) and our backend uses a giant model~\cite{lin2025depthanything3}; \textbf{Ours~($\Omega$)} denotes the same pipeline using VGGT-$\Omega$~\cite{wang2026vggtomega} as the backend, which we include to show that our system is agnostic to the underlying geometric foundation model. Unless stated otherwise, \textbf{Ours} represents the monocular system.

\subsection{Evaluation}

\noindent
\textbf{Large-scale outdoor scenes.} At the kilometer scale, \papername{} is the only online method that matches or surpasses offline global reconstruction methods (Tab.~\ref{tab:kitti_pose}--\ref{tbl:spires}). On KITTI (Tab.~\ref{tab:kitti_pose}), our monocular variant achieves 13.11\,m average ATE, outperforming the leading online competitor (LoGeR, 18.65\,m) by 30\%, the best chunk-based method (DA3-Long, 16.83\,m) by 22\%, and slightly outperforming offline SfM (Glob3R~\cite{deng2026glob3r}, 13.21\,m).  Our advantage becomes larger when evaluating on more diverse outdoor trajectories. 
\begin{table}[t]
\centering
\setlength{\tabcolsep}{1pt}
\colorlet{secolor}{lightgray}
\newcommand{\se}{\color{secolor}}
\caption{\textbf{Comparison of ATE [m]$\downarrow$ on VBR.}}
\vspace{-2mm}
\resizebox{\columnwidth}{!}{%
\begin{tabular}{c@{\hspace{6pt}}l ccccccc c}
\toprule
& \multirow{2}{*}{\textbf{Method}} 
& \textbf{col\_0} & \textbf{cam\_0} & \textbf{cam\_1} & \textbf{pin\_0} & \textbf{spa\_0} & \textbf{dia\_0} & \textbf{cia\_1} & \multirow{2}{*}{\textbf{Avg.}} \\
& 
& \scriptsize 8.8k/1.5km & \scriptsize 12.0k/2.7km & \scriptsize 11.7k/3.0km & \scriptsize 11.1k/1.3km & \scriptsize 14.1k/1.6km & \scriptsize 10.0k/1.0km & \scriptsize 18.8k/5.2km & \\
\midrule
\multirow{3}{*}{\rotatebox{90}{\textbf{Offline}}}
& VGGT-Long~\citep{deng2025vggtlong}   & 45.73             & 132.68            & 115.65            & 64.40             & 58.49             & 33.66             & 187.97            & 91.23 \\
& $\pi$3-Long~\cite{wang2025pi3}       & 81.63             & 86.54             & 71.49             & 49.85             & 54.27             & 28.07             & 118.81            & 70.09 \\
& DA3-Long~\cite{lin2025depthanything3}& 65.78             & 88.12             & 102.28            & 41.08             & 55.93             & 4.15              & \underline{11.21} & 52.65 \\
\midrule
\multirow{6}{*}{\rotatebox{90}{\textbf{Online}}}
& VGGT-SLAM~\citep{maggio2025vggtslam} & 102.92            & 110.77            & 89.69             & 72.98             & 62.67             & 35.62             & 144.02            & 88.38 \\
& DA3-SLAM                             & 97.19             & 93.57             & 91.61             & 81.82             & 65.44             & 36.61             & 192.95            & 94.17 \\
& LoGeR~\cite{zhang2026loger}          & 49.38             & 22.44             & 34.90             & 11.08             & 27.05             & 32.88             & 44.51             & 31.75 \\
& LingBot-Map~\citep{chen2026lingbotmap}& 15.02            & \underline{20.34} & 9.94              & 30.60             & 23.59             & 22.75             & 64.34             & 26.65 \\
& \textbf{Ours ($\Omega$)}             & \textbf{2.52}     & 26.43             & \textbf{3.72}     & \textbf{3.95}     & \textbf{5.10}     & \textbf{2.66}     & \textbf{5.66}     & \textbf{7.15} \\
& \textbf{Ours}                        & \underline{3.95}  & \textbf{9.29}     & \underline{6.73}  & \underline{10.75} & \underline{5.94}  & \underline{3.36}  & 11.92             & \underline{7.42} \\
\midrule
\multirow{4}{*}{\rotatebox{90}{\textbf{LiDAR}}}
& PIN-SLAM~\cite{pan2024pinslam}       & 1.69 & \textbf{0.74} & \textbf{0.14} & 0.46 & 0.31 & 1.10 & 1.24 & 0.81 \\
& \quad \se$\hookrightarrow$ $\mathrm{SE}(3)$
                                       & \se 1.70 & \se\textbf{0.74} & \se\textbf{0.17} & \se 0.50 & \se 0.32 & \se 1.10 & \se 1.25 & \se 0.83 \\
& \textbf{Ours (LiDAR)}                & \textbf{0.22} & 1.01 & 0.28 & \textbf{0.18} & \textbf{0.10} & \textbf{0.25} & \textbf{0.50} & \textbf{0.36} \\
& \quad \se$\hookrightarrow$ $\mathrm{SE}(3)$
                                       & \se\textbf{0.22} & \se 1.02 & \se 0.28 & \se\textbf{0.37} & \se\textbf{0.10} & \se\textbf{0.25} & \se\textbf{0.52} & \se\textbf{0.39} \\
\bottomrule

\end{tabular}%
}
\label{tbl:vbr}
\vspace{-1mm}
\end{table}
On VBR (Tab.~\ref{tbl:vbr}), with urban sequences reaching up to 18.8k frames and 5.2\,km, \papername{} reduces average ATE from 26.65\,m (LingBot-Map) and 31.75\,m (LoGeR) to 7.42\,m (7.15\,m with $\Omega$), achieving a 72\% error reduction, and is also significantly better than all chunk-based methods.  On Oxford Spires (Tab.~\ref{tbl:spires} and Fig.~\ref{fig:qual}), \papername{} reduces average ATE from 8.59\,m (LoGeR) to 2.32\,m (2.08\,m with $\Omega$), a 73\% reduction. Our method shows significantly better camera tracking performance on Christ Church 5, which contains challenging indoor-to-outdoor-to-indoor transitions, demonstrating its robustness across different regimes.

\noindent
\textbf{Building-scale indoor-outdoor capture.} Following VidMap~\cite{pataki2026vidmap}, we evaluate on LaMAR~\cite{sarlin2022lamar} and CroCoDL~\cite{Blum2025crocodl} in Tab.~\ref{tab:lamar_crocodl_pose_auc}, both of which contain trajectories averaging over 100\,m with challenging conditions and out-of-distribution environments. Compared to existing SLAM systems, such as DA3-SLAM, the advantage of our method grows with window length. On CroCoDL, a dataset with video captured in out-of-domain disaster-site buildings, our method outperforms all online methods by a significant margin, and surpasses offline GLOMAP (76.4\%) and DA3-Long (88.9\%). These results demonstrate the effectiveness of our system for preventing drift under severe out-of-domain conditions.

\begin{table}[tb]
\centering
\caption{\textbf{Evaluation on the LaMAR and CroCoDL (Uncalibrated).} We report the window AUC of the translation error~\cite{pataki2026vidmap}.}
\vspace{-2mm}
\label{tab:lamar_crocodl_pose_auc}
\setlength{\tabcolsep}{2pt}
\resizebox{\columnwidth}{!}{%
\begin{tabular}{cl ccccc ccccc}
\toprule
& \multirow{2}{*}{\textbf{Method}} & \multicolumn{5}{c}{\textbf{LaMAR~\cite{sarlin2022lamar}}} & \multicolumn{5}{c}{\textbf{CroCoDL~\cite{Blum2025crocodl}}} \\
\cmidrule(lr){3-7} \cmidrule(lr){8-12}
& & 10\,m & 25\,m & 50\,m & 100\,m & Full & 10\,m & 25\,m & 50\,m & 100\,m & Full \\
\midrule
\multirow{3}{*}{\rotatebox{90}{\textbf{Offline}}}
& GLOMAP~\cite{pan2024glomap}          & 76.4 & 69.4 & 60.1 & 53.6 & 62.3 & 77.8 & 66.6 & 60.0 & 65.9 & 76.4 \\
& DA3-Long~\cite{lin2025depthanything3} & 86.7 & 86.7 & 84.6 & 78.1 & 76.5 & 86.2 & 88.1 & 88.5 & 88.8 & 88.9 \\
& VidMap~\cite{pataki2026vidmap}        & \textbf{91.9} & \textbf{92.3} & \textbf{89.9} & \textbf{89.3} & \textbf{88.5} & \textbf{94.0} & \textbf{95.0} & \textbf{95.3} & \textbf{93.0} & \textbf{95.5} \\
\midrule
\multirow{9}{*}{\rotatebox{90}{\textbf{Online}}}
& LoGeR~\cite{zhang2026loger}          & 65.9 & 64.5 & 60.3 & 60.1 & 59.8 & 50.0 & 56.7 & 57.9 & 63.7 & 63.8 \\
& LingBot-Map~\cite{chen2026lingbotmap} & 69.2 & 75.6 & 75.0 & 72.7 & 69.6 & 54.4 & 66.3 & 69.9 & 72.2 & 72.0 \\
& VGGT-SLAM2~\cite{maggio2026vggtslam2} & 74.0 & 69.2 & 60.3 & 49.4 & 56.9 & 62.8 & 64.3 & 66.0 & 62.5 & 74.9 \\
& DA3-SLAM                             & 82.5 & 76.4 & 67.9 & 60.8 & 62.5 & 77.1 & 72.5 & 65.9 & 57.6 & 70.9 \\
& MASt3R-SLAM~\cite{murai2025mast3rslam}& 60.8 & 63.6 & 55.6 & 47.3 & 51.5 & 51.8 & 57.3 & 60.2 & 65.2 & 70.0 \\
& MegaSaM~\cite{li2025megasam}         & 77.5 & 68.4 & 59.9 & 51.4 & 52.7 & 50.2 & 50.2 & 43.5 & 35.2 & 50.1 \\
& ViPE~\cite{huang2025vipe}            & 86.6 & 83.8 & 80.2 & 78.1 & 76.6 & 66.7 & 68.2 & 68.6 & 70.7 & 71.4 \\
& \textbf{Ours ($\Omega$)}             & \textbf{88.9} & \textbf{89.3} & \textbf{87.7} & \textbf{85.5} & \textbf{84.7} & 88.6 & 90.1 & \textbf{91.0} & \textbf{92.2} & \textbf{91.7} \\
& \textbf{Ours}                        & 88.7 & 87.2 & 83.8 & 78.0 & 75.7 & \textbf{90.9} & \textbf{90.8} & 90.1 & 91.8 & 91.2 \\
\bottomrule
\end{tabular}%
\vspace{-1mm}
}

\end{table}

\begin{table}[t]
\centering
\caption{\textbf{Comparison of ATE [cm]$\downarrow$ on TUM RGB-D.} $^\dagger$ denotes methods that evaluate on keyframe poses only.}
\vspace{-2mm}
\label{tab:tum_pose}
\scriptsize
\setlength{\tabcolsep}{1.5pt}
\resizebox{\columnwidth}{!}{
\begin{tabular}{clccccccccc c}
\toprule
&\textbf{Method} & 360 & desk & desk2 & floor & plant & room & rpy & teddy & xyz & \textbf{Avg.} \\

\midrule
\multirow{3}{*}{\rotatebox{90}{\textbf{Offline}}}
& Scal3R \cite{xie2026scal3r}
& 6.6 & 5.5 & 2.9 & 18.9 & 4.5 & 11.0 & 3.2 & 9.5 & 5.0 & 7.4 \\
& Glob3R~\cite{deng2026glob3r} 
& 8.1 & 1.8 & 2.6 & 3.0 & \textbf{1.8} & 5.0 & 2.2 & 3.7 & 0.9 & 3.2 \\
& SAILRecon \citep{deng2025sail}
& 7.0 & 2.4 & 4.2 & 10.7 & 3.1 & 11.3 & 2.0 & 3.7 & 1.2 & 5.1 \\

\midrule
\multirow{3}{*}{\rotatebox{90}{\textbf{Chunk}}}
& VGGT-Long \citep{deng2025vggtlong}
& 11.8 & 5.8 & 11.1 & 11.8 & 7.1 & 15.5 & 14.0 & 12.0 & 9.9 & 11.0 \\
& $\pi$3-Long~\cite{wang2025pi3}
& 11.5 & 4.7 & 5.2 & 16.0 & 8.5 & 11.4 & 14.3 & 8.1 & 5.2 & 9.4 \\
& DA3-Long~\cite{lin2025depthanything3}
& 5.9 & 3.4 & 4.2 & 10.7 & 6.0 & 10.5 & 20.6 & 12.6 & 4.4 & 8.7 \\

\midrule
\multirow{9}{*}{\rotatebox{90}{\textbf{Online}}}
& DROID-SLAM$^\dagger$ \citep{teed2021droidslam}
& 20.2 & 3.2 & 9.1 & 6.4 & 4.5 & 91.8 & 5.6 & 4.5 & 1.2 & 15.8 \\
& MASt3R-SLAM$^\dagger$ \citep{murai2025mast3rslam}
& 7.0 & 3.5 & 5.5 & 5.6 & 3.5 & 11.8 & 4.1 & 11.4 & 2.0 & 6.0 \\
& LoGeR~\cite{zhang2026loger}
& 10.6 & 3.2 & 4.6 & 10.3 & 4.6 & 9.8 & 3.0 & 8.4 & 2.2 & 6.3 \\
& MUSt3R~\cite{cabon2025must3r}
& 7.8 & 5.1 & 7.1 & 5.0 & 4.0 & 9.9 & 4.3 & 4.2 & 1.3 & 5.4 \\
& VGGT-SLAM2$^\dagger$ \citep{maggio2026vggtslam2}
& 5.0 & 2.5 & 2.9 & 10.2 & 2.6 & 6.3 & 2.6 & 3.8 & 1.4 & 4.1 \\
& LingBot-Map \citep{chen2026lingbotmap}
& 6.3 & 2.8 & 4.4 & 5.9 & 4.4 & 9.2 & 2.4 & 3.6 & 1.1 & 4.4 \\
& AMB3R-VO \citep{wang2026amb3r}
& 4.6 & 1.9 & 2.8 & 3.2 & 2.9 & 5.8 & 2.3 & 3.7 & 1.1 & 3.2 \\
& \textbf{Ours (Mono)}
& \textbf{3.6} & \textbf{1.4} & \textbf{2.2} & \textbf{2.6} & 2.3 & 2.9 & \textbf{1.8} & 2.9 & \textbf{0.7} & 2.3 \\
& \textbf{Ours (RGB-D)}
& \textbf{3.6} & \textbf{1.4} & \textbf{2.2} & 2.7 & 2.0 & \textbf{2.8} & \textbf{1.8} & \textbf{2.8} & 0.8 & \textbf{2.2} \\

\bottomrule
\end{tabular}
\vspace{-1mm}
}
\end{table}
\begin{table}[tb]
\centering
\caption{\textbf{Evaluation on the ETH3D-SLAM and EuRoC (Uncalibrated).} We report AUC of the translation error~\cite{pataki2026vidmap}.}
\vspace{-2mm}
\label{tab:eth3d_euroc_pose_auc}
\setlength{\tabcolsep}{2.5pt}
\resizebox{\columnwidth}{!}{%
\begin{tabular}{cl ccccc ccccc}
\toprule
& \multirow{2}{*}{\textbf{Method}} & \multicolumn{5}{c}{\textbf{ETH3D-SLAM~\cite{schops2019badslam}}} & \multicolumn{5}{c}{\textbf{EuRoC~\cite{Burri2016euroc}}} \\
\cmidrule(lr){3-7} \cmidrule(lr){8-12}
& & 5\,cm & 10\,cm & 50\,cm & 1\,m & 10\,m & 5\,cm & 10\,cm & 50\,cm & 1\,m & 10\,m \\
\midrule
\multirow{3}{*}{\rotatebox{90}{\textbf{Offline}}}
& GLOMAP~\cite{pan2024glomap}   & 20.4 & 30.9 & 58.7 & 67.4 & 77.8 & 0.4  & 1.8  & 16.1 & 29.7 & 70.7 \\
& DA3-Long~\cite{lin2025depthanything3}    & 19.0 & 37.2 & 73.6 & 84.4 & 98.0 & 0.1  & 0.8  & 31.4 & 57.6 & 95.4 \\
& VidMap~\cite{pataki2026vidmap}      & \textbf{51.4} & \textbf{68.3} & \textbf{90.1} & \textbf{94.0} & \textbf{99.0} & \textbf{11.5} & \textbf{31.6} & \textbf{79.1} & \textbf{89.4} & \textbf{99.0} \\
\midrule
\multirow{8}{*}{\rotatebox{90}{\textbf{Online}}}
& LoGeR~\cite{zhang2026loger}       & 6.4  & 16.9 & 57.7 & 73.8 & 95.9 & 0.1  & 0.5  & 18.2 & 45.1 & 92.6 \\
& LingBot-Map~\cite{chen2026lingbotmap} & 13.1 & 27.8 & 68.3 & 81.3 & 97.2 & 0.1  & 0.7  & 32.0 & 55.5 & 91.7 \\
& VGGT-SLAM2~\cite{maggio2026vggtslam2}  & 26.4 & 44.1 & 75.7 & 84.3 & 97.1 & 1.3  & 7.8  & 55.4 & 74.2 & 96.9 \\
& MASt3R-SLAM~\cite{murai2025mast3rslam} & 8.9  & 21.6 & 65.3 & 77.0 & 96.4 & 1.6  & 5.6  & 51.0 & 71.9 & 97.0 \\
& MegaSaM~\cite{li2025megasam}     & 38.4 & 53.9 & 79.2 & 86.7 & 98.0 & 0.8  & 5.7  & 50.3 & 73.2 & 97.3 \\
& ViPE~\cite{huang2025vipe}        & 38.7 & 54.9 & 78.3 & 85.7 & 97.6 & 1.3  & 6.2  & 55.4 & 75.0 & 94.0 \\
& \textbf{Ours (Mono)}  & 44.0 & 60.6 & 84.7 & 90.4 & 98.8 & \textbf{20.3} & \textbf{44.0} & \textbf{85.5} & \textbf{92.8} & \textbf{99.3} \\
& \textbf{Ours (RGB-D)} & \textbf{44.8} & \textbf{61.8} & \textbf{85.3} & \textbf{90.7} & \textbf{98.9} & -- & -- & -- & -- & -- \\
\bottomrule
\end{tabular}%
}
\vspace{-1mm}
\end{table}

\noindent
\textbf{Indoor scenes.} Room-scale sequences test the opposite regime, where cumulative drift is minimal, and tracking quality is primarily determined by local accuracy. Our method also leads consistently in this case. On TUM RGB-D (Tab.~\ref{tab:tum_pose}), \papername{} achieves a 2.3\,cm average ATE, a 28\% error reduction over the previous best online method, AMB3R-VO (3.2\,cm). Notably, we also surpass every offline and chunk-based method, despite using a significantly smaller frame budget for our base model to retain real-time performance. This shows our system does not trade local fidelity for global consistency. On ETH3D-SLAM (Tab.~\ref{tab:eth3d_euroc_pose_auc}), \papername{} improves AUC@5\,cm from 38.7\% (ViPE) to 44.0\% and remains ahead at every threshold. On EuRoC, where the fast drone motion causes all prior online systems to collapse at tight thresholds, our method remains robust and even outperforms the offline VidMap (11.5) despite its extensive optimization.

\begin{table}[t]
\centering
\caption{\textbf{Comparison of ATE [cm]$\downarrow$ on Bonn Dynamic Dataset.}}
\vspace{-2mm}
\label{tab:bonn_tracking}
\setlength{\tabcolsep}{1.5pt}
\resizebox{\columnwidth}{!}{%
\begin{tabular}{lccccccccc}
\toprule
\textbf{Method} & \textbf{bal\_1} & \textbf{bal\_2} & \textbf{crd\_1} & \textbf{crd\_2} & \textbf{per\_1} & \textbf{per\_2} & \textbf{mov\_1} & \textbf{mov\_2} & \textbf{Avg.} \\
\midrule
DSO~\cite{engel2017dso}                   & 7.3  & 21.8 & 10.1 & 7.6  & 30.6 & 26.5 & 4.7  & 11.2 & 15.0 \\
DROID-SLAM~\cite{teed2021droidslam}       & 7.5  & 4.1  & 5.2  & 6.5  & 4.3  & 5.4  & 2.3  & 4.0  & 4.9  \\
MonoGS~\cite{matsuki2024monogs}           & 15.3 & 17.3 & 11.3 & 7.3  & 26.4 & 35.2 & 22.2 & 47.2 & 22.8 \\
DynaMoN~\cite{schischka2024dynamon}       & 6.8  & 3.8  & 6.1  & 5.6  & 2.4  & 3.5  & 1.4  & 2.6  & 4.0  \\
MonST3R~\cite{zhang2025monst3r}           & 5.4  & 7.2  & 5.4  & 6.9  & 11.9 & 11.1 & 3.3  & 7.4  & 7.3  \\
MegaSaM~\cite{li2025megasam}              & 3.7  & 2.6  & 1.6  & 7.2  & 4.1  & 4.0  & 1.4  & 3.4  & 3.5  \\
WildGS-SLAM~\cite{zheng2025wildgsslam}    & 2.8  & 2.4  & 1.5  & 2.3  & 3.1  & 2.7  & 1.6  & 2.2  & 2.3  \\
\textbf{Ours ($\Omega$)}                  & 1.7  & 2.0  & 1.3  & 1.3  & 3.6  & 2.2  & 1.6  & 2.7  & 2.1  \\
\textbf{Ours}                             & \textbf{1.3} & \textbf{1.4} & \textbf{1.1} & \textbf{1.2} & \textbf{1.8} & \textbf{1.3} & \textbf{1.0} & \textbf{1.1} & \textbf{1.3} \\
\bottomrule
\end{tabular}%
}
\vspace{-1mm}
\end{table}

\noindent
\textbf{Dynamic scenes.}
We additionally show the performance on the Bonn dynamic dataset (Tab.~\ref{tab:bonn_tracking}). \papername{} achieves a 1.3\,cm average ATE and is the most accurate method on all 8 sequences, reducing the error of WildGS-SLAM (2.3\,cm) by 43\% and of MegaSaM (3.5\,cm) by 63\%. Notably, these results are obtained using the exact same pipeline for static scenes, without any task-specific adaptation. This is because our system performs graph optimization over relative camera poses rather than bundle adjustment, thereby avoiding the need to filter out moving objects or model complex dynamic correspondences.

\noindent
\textbf{Additional modalities.} Beyond monocular tracking, \papername{} can also integrate stereo and RGB-D sensors by fixing the scale ($s_{ij} = 1.0$) of our pose graph, restricting the optimization from $\mathrm{Sim}(3)$ to $\mathrm{SE}(3)$ and alleviating scale drift. 
On KITTI (Tab.~\ref{tab:kitti_pose}), incorporating stereo sensors reduces the average ATE to 12.15\,m. In indoor scenes, RGB-D input improves TUM (Tab.~\ref{tab:tum_pose}) ATE to 2.2\,cm and ETH3D-SLAM (Tab.~\ref{tab:eth3d_euroc_pose_auc}) AUC@5\,cm to 44.8\%. 

These gains are relatively modest due to the range limitations and noise of depth sensors. In contrast, with LiDAR as an additional input, the performance of \papername{} improves substantially. On KITTI (Tab.~\ref{tab:kitti_pose}), \papername{} (LiDAR) reduces average ATE to 0.95\,m in $\mathrm{Sim}(3)$ (1.01\,m in $\mathrm{SE}(3)$), outperforming PIN-SLAM (1.22\,m), a state-of-the-art LiDAR SLAM based on neural implicit representations. 
On KITTI-01, a 2.5\,km highway sequence with limited visual features and rapid camera motion where monocular tracking drifts severely, the LiDAR input reduces ATE from 55.91\,m to 2.95\,m. On VBR (Tab.~\ref{tbl:vbr}) with more diverse camera motions, \papername{} (LiDAR) reduces average ATE to 0.36\,m (0.39\,m in $\mathrm{SE}(3)$).

\subsection{Analysis}

\noindent
\textbf{Computation analysis.} As shown in Tab.~\ref{tab:runtime}, \papername{} operates in real time on a single NVIDIA RTX 4090, achieving 10.2--17.6\,FPS in monocular mode and 31.3--47.8\,FPS with LiDAR, while keeping peak memory around 10--14\,GB without growing with sequence length. The inference of the geometric foundation models accounts for most of the per-frame latency, whereas loop retrieval and $\mathrm{Sim}(3)$ graph optimization incur negligible cost. By optimizing relative poses between submaps rather than bundle adjustment over millions of 3D points, our hierarchical backend can easily scale to multi-kilometer sequences.

\begin{table}[t]
\centering
\caption{\textbf{Runtime and memory of \papername{}.} We additionally report the cost decomposition of mapping, loop retrieval, and the pose graph optimization.}
\vspace{-2mm}
\label{tab:runtime}
\setlength{\tabcolsep}{4pt}
\resizebox{\columnwidth}{!}{%
\begin{tabular}{ll cc ccc}
\toprule
\multirow{2}{*}{\textbf{Method}} & \multirow{2}{*}{\textbf{Dataset}}
& \multirow{2}{*}{\textbf{FPS}$\uparrow$} & \multirow{2}{*}{\textbf{Mem.\ [GB]}$\downarrow$}
& \multicolumn{3}{c}{\textbf{Time [ms/frame]}$\downarrow$} \\
\cmidrule(lr){5-7}
& & & & Mapping & Retrieval & PGO \\
\midrule
Mono          & KITTI & 17.6 & 10.3 & 56.5 & 0.7 & 0.17 \\
Mono          & VBR   & 10.2 & 14.0 & 97.5 & 1.0 & 0.12 \\
LiDAR         & KITTI & 47.8 & 10.5 & 20.8 & 0.7 & 0.05 \\
LiDAR         & VBR   & 31.3 & 14.1 & 44.0 & 1.0 & 0.04 \\

\bottomrule
\end{tabular}%
}
\vspace{-1mm}
\end{table}

\begin{table}[t]
\centering
\caption{\textbf{Ablation of \papername{} (Mono) on VBR.} We report W-AUC [\%]$\uparrow$ over all 7 sequences.}
\vspace{-2mm}
\label{tab:vbr_ablation}
\setlength{\tabcolsep}{4pt}
\resizebox{\columnwidth}{!}{%
\begin{tabular}{l cccccc}
\toprule
\multirow{2}{*}{\textbf{Variant}}
& \multicolumn{6}{c}{\textbf{W-AUC [\%]}$\uparrow$} \\
\cmidrule(lr){2-7}
& \textbf{10\,m} & \textbf{50\,m} & \textbf{100\,m} & \textbf{500\,m} & \textbf{1000\,m} & \textbf{full} \\
\midrule
w/o LC               & \textbf{89.38} & 86.31          & 80.94          & 55.39          & 47.82          & 42.02          \\
w/o whitening        & 88.11          & 84.65          & 80.92          & 77.47          & 82.79          & 86.97          \\
w/o SLCM              & 89.15          & 86.71          & 84.28          & 84.36          & 88.35          & \textbf{90.46} \\
\textbf{Ours (Full)} & 89.14          & \textbf{87.81} & \textbf{85.57} & \textbf{86.00} & \textbf{90.17} & 90.39          \\
\bottomrule
\end{tabular}%
}
\vspace{-1mm}
\end{table}

\noindent
\textbf{Ablation study.} We evaluate various components of our backend on VBR in Tab.~\ref{tab:vbr_ablation}. 
Without loop closure (\emph{w/o LC}), \papername{} maintains reliable camera tracking up to 100\,m, but eventually drifts over time, dropping W-AUC to 55.39\% at 500\,m and 42.02\% across full trajectories. 
Removing the whitening matrix (\emph{w/o whitening}) degrades accuracy across all windows, as unnormalized translation residuals overpower pose graph optimization over longer baselines. 
Without sparse long-context mapping (\emph{w/o SLCM}), camera tracking degrades across 50--1000\,m where loop closures are sparse or absent. 
This shows that our long-context edges provide effective mid-range constraints that bound drift before loop closure triggers.

In Tab.~\ref{tab:lidar_ablation}, we ablate the effect of geometric foundation models in our LiDAR pipeline. Since LiDAR with ICP-based methods is quite accurate, the geometric foundation model here acts as a good prior to initialize registration during loop closure. This reduces average ATE from 1.45\,m to 1.23\,m by avoiding local geometric minima under viewpoint changes.

\begin{table}[t]
\centering
\caption{\textbf{Ablation of \papername{} (LiDAR) on KITTI.} We report ATE [m]$\downarrow$ averaged over 3 sequences with loop closure.}
\vspace{-1mm}
\label{tab:lidar_ablation}
\setlength{\tabcolsep}{6pt}
\resizebox{0.65\columnwidth}{!}{%
\begin{tabular}{lcc}
\toprule
\textbf{Variant} & \textbf{ATE} & \textbf{ATE ($\mathrm{SE}(3)$)} \\
\midrule
w/o GFM init         & 1.39          & 1.45          \\
\textbf{Ours (Full)} & \textbf{1.17} & \textbf{1.23} \\
\bottomrule
\end{tabular}%
}
\vspace{-1mm}
\end{table}

\noindent
\textbf{Limitations and future work.}
Currently, \papername{} represents 3D structure through point clouds rather than a unified, compact surface representation. 
As a result, our map may exhibit duplicated surfaces or ghosting artifacts in perceptually ambiguous regions. Integrating compact representations, such as neural implicit fields, could help to maintain a globally consistent map. 
Additionally, while \papername{} effectively incorporates stereo, RGB-D, and LiDAR at the system level via graph optimization, the underlying geometric foundation models operate purely on RGB inputs without conditioning on those additional modalities. Exploring foundation models with native multi-modal conditioning is a promising direction to further enhance geometric fidelity and tracking accuracy.

\section{Conclusion}

We have presented \papername{}, a real-time monocular SLAM system capable of kilometer-scale tracking over 10k frames on a single consumer-grade GPU. By coupling a lightweight front-end with a hierarchical backend, our system progressively enforces local, mid-level, and global consistency, effectively scaling to large-scale reconstruction. Since our system does not rely on bundle adjustment with a static world assumption, we show that it can deal with complex dynamic environments out of the box. Furthermore, we have shown that our system can be extended to additional modalities, including stereo, RGB-D, and LiDAR, to further improve its robustness towards various challenging environments.

\noindent
\textbf{Acknowledgments.} The research presented here has been supported by a sponsored research award from Cisco Research and the UCL Centre for Doctoral Training in Foundational AI under UKRI grant number EP/S021566/1.

{
    \small
    \bibliographystyle{ieeenat_fullname}
    \bibliography{main}
}

\end{document}